\documentclass[sigconf, nonacm]{acmart}

\newcommand\vldbdoi{10.14778/3565838.3565859}
\newcommand\vldbpages{4093 - 4105}
\newcommand\vldbvolume{15}
\newcommand\vldbissue{13}
\newcommand\vldbyear{2022}
\newcommand\vldbauthors{\authors}
\newcommand\vldbtitle{\shorttitle} 
\newcommand\vldbavailabilityurl{https://github.com/ChengYuHsieh/Nemo}
\newcommand\vldbpagestyle{empty}

\usepackage{amsmath,amsfonts,bm}

\def\eqref#1{Eq.~\ref{#1}}

\def\1{\bm{1}}

\DeclareMathAlphabet{\mathsfit}{\encodingdefault}{\sfdefault}{m}{sl}
\SetMathAlphabet{\mathsfit}{bold}{\encodingdefault}{\sfdefault}{bx}{n}

\newcommand{\E}{\mathbb{E}}

\DeclareMathOperator*{\argmax}{arg\,max}

\usepackage{enumitem}
\usepackage{algorithm}
\usepackage{algorithmic}
\usepackage[ruled,linesnumbered,algo2e]{algorithm2e}
\usepackage{multirow}
\usepackage{adjustbox}
\usepackage{subcaption}
\usepackage{tabularx}
\newcolumntype{C}{>{\centering\arraybackslash}X}
\usepackage{booktabs, array}
\usepackage{dsfont}
\usepackage{bbm}
\usepackage{pifont}

\begin{document}
\title[Nemo: Guiding and Contextualizing Weak Supervision for Interactive Data Programming]{Nemo: Guiding and Contextualizing Weak Supervision\\for Interactive Data Programming}

\author{Cheng-Yu Hsieh}
\affiliation{%
  \institution{University of Washington}
}
\email{cydhsieh@cs.washington.edu}

\author{Jieyu Zhang}
\affiliation{%
  \institution{University of Washington}
}
\email{jieyuz2@cs.washington.edu}

\author{Alexander Ratner}
\affiliation{%
  \institution{University of Washington}
}
\affiliation{%
  \institution{Snorkel AI, Inc.}
}
\email{ajratner@cs.washington.edu}

\begin{abstract}
Weak Supervision (WS) techniques allow users to efficiently create large training datasets by programmatically labeling data with heuristic sources of supervision. 
While the success of WS relies heavily on the provided labeling heuristics, the process of how these heuristics are created in practice has remained under-explored.
In this work, we formalize the development process of labeling heuristics as an interactive procedure, built around the existing workflow where users draw ideas from a selected set of \textit{development data} for designing the heuristic sources. With the formalism, shown in Figure 1, we study two core problems of (1) how to strategically select the development data to \textit{guide} users in efficiently creating informative heuristics, and (2) how to exploit the information within the development process to \textit{contextualize} and better learn from the resultant heuristics.
Building upon two novel methodologies that effectively tackle the respective problems considered, we present Nemo, an end-to-end interactive system that
improves the overall productivity of WS learning pipeline by an average 20\% (and up to 47\% in one task) compared to the prevailing WS approach.
\end{abstract}

\maketitle

\pagestyle{\vldbpagestyle}
\begingroup\small\noindent\raggedright\textbf{PVLDB Reference Format:}\\
\vldbauthors. \vldbtitle. PVLDB, \vldbvolume(\vldbissue): \vldbpages, \vldbyear.\\
\href{https://doi.org/\vldbdoi}{doi:\vldbdoi}
\endgroup
\begingroup
\renewcommand\thefootnote{}\footnote{\noindent
This work is licensed under the Creative Commons BY-NC-ND 4.0 International License. Visit \url{https://creativecommons.org/licenses/by-nc-nd/4.0/} to view a copy of this license. For any use beyond those covered by this license, obtain permission by emailing \href{mailto:info@vldb.org}{info@vldb.org}. Copyright is held by the owner/author(s). Publication rights licensed to the VLDB Endowment. \\
\raggedright Proceedings of the VLDB Endowment, Vol. \vldbvolume, No. \vldbissue\ %
ISSN 2150-8097. \\
\href{https://doi.org/\vldbdoi}{doi:\vldbdoi} \\
}\addtocounter{footnote}{-1}\endgroup

\ifdefempty{\vldbavailabilityurl}{}{
\vspace{.3cm}
\begingroup\small\noindent\raggedright\textbf{PVLDB Artifact Availability:}\\
The source code, data, and/or other artifacts have been made available at \url{\vldbavailabilityurl}.
\endgroup
}

\section{Introduction}
Manual labeling and curation of training datasets has increasingly become one of the major bottlenecks when deploying modern machine learning models in practice.
In response, recent weak supervision approaches, wherein cheaper but noisier forms of labels are used, have received increasing research and industry attention.
In one recent paradigm for weak supervision, data programming (DP) \citep{ratner2016data}, users are able to quickly create and manage large training datasets by encoding domain knowledge and labeling heuristics into a set of \textit{labeling functions} (LFs). Each of these functions, serving as a weak supervision source, programmatically annotates a subset of data points with possibly noisy weak labels. As the LFs may have varying accuracies, different LFs could suggest conflicting votes on certain data points. As a result, researchers have developed various modeling techniques to denoise and aggregate the weak labels, provided by different weak supervision sources, into probabilistic training labels that can be utilized to train downstream models~\citep{ratner2016data,ratner2019training, fu2020fast}.
Despite its recent emergence, the DP paradigm has powered many industrial-scale systems across various domains~\citep{ratner2017snorkel,bach2019snorkel,Bringer2019OspreyWS,re2019overton}.


\begin{figure}
    \centering
    \includegraphics[width=\linewidth]{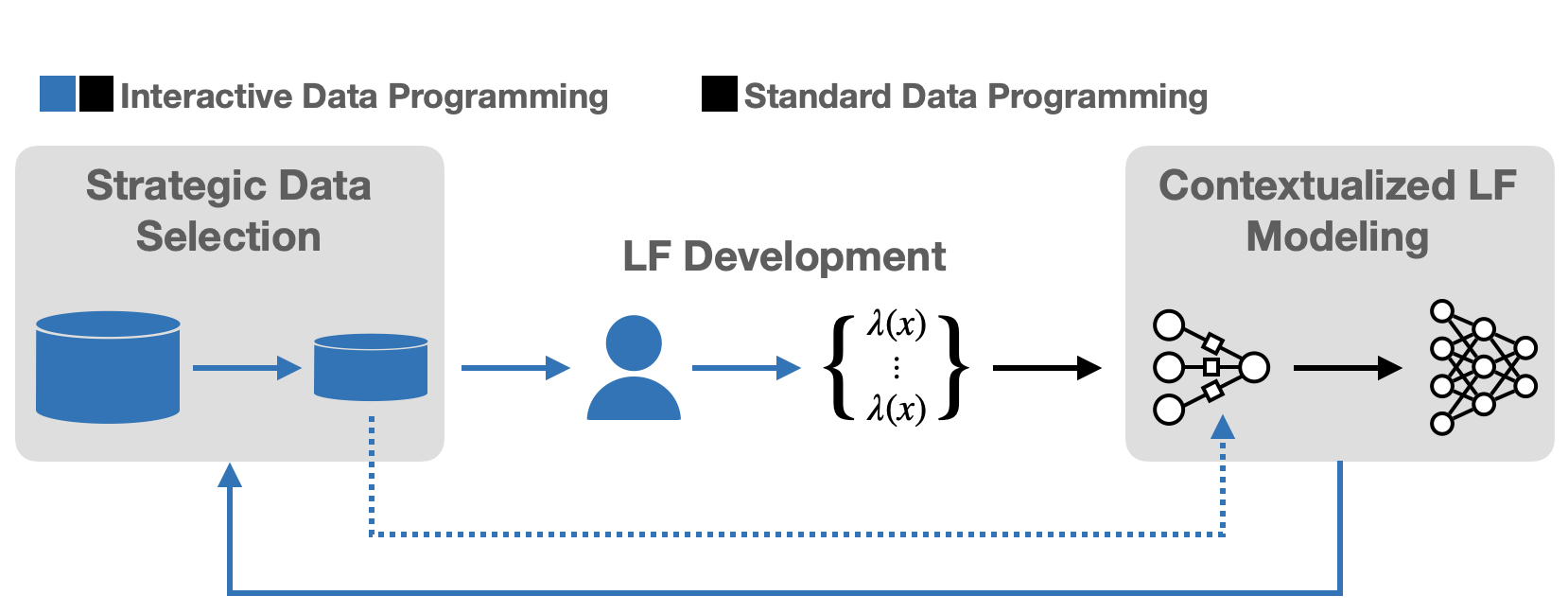}
    \vspace{-5mm}
    \caption{Interactive Data Programming (IDP) considers the entire data programming (DP) pipeline as an iterative cycle with two core problems of (1) strategic data selection and (2) contextualized LF modeling. The blue-colored components/steps in IDP are typically neglected in existing DP pipeline (the black-colored sub-procedures) focusing more exclusively on modeling and learning from a \textit{given} set of LFs.}
    \label{fig:idp}
    \vspace{-5mm}
\end{figure}

\begin{figure}[!t]
    \centering
    \includegraphics[width=\linewidth]{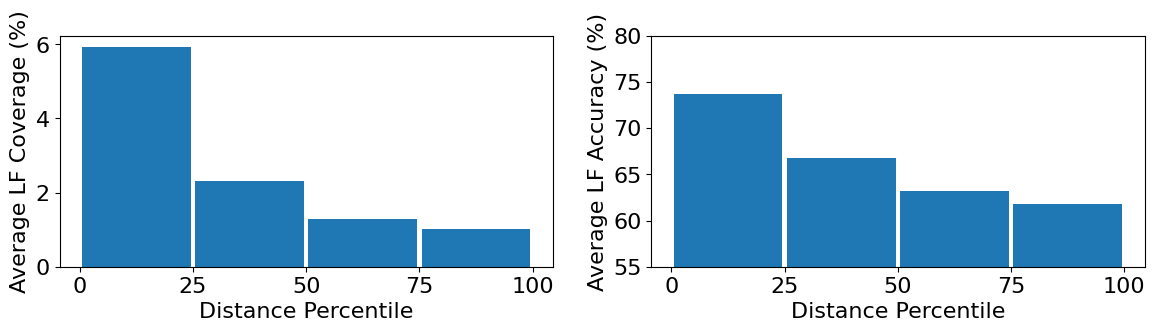}
    \vspace{-8mm}
    \caption{LFs generally have higher coverage (left) and accuracy (right) on the data subspace within closer proximity to their development data. For each LF, we organize all examples into 4 subspaces based on the percentile of their distance to the development data, and compute the LF’s coverage/accuracy in each data subspaces. The plots are averaged results over 100 LFs on Amazon Review Dataset.}
    \label{fig:lf_acc_quantile}
\end{figure}

In leveraging the DP paradigm, the crucial---yet understudied---first step is to develop a set of LFs as the weak supervision sources.
In practice, users typically look at a small set of data samples selected from the unlabeled dataset, called the \textit{development data}, to draw ideas for writing LFs~\citep{ratner2017snorkel}.
As a result of this common workflow, we especially note that the LFs developed are directly affected by and biased with respect to the data seen by the users during the development process.
Particularly, an LF created with heuristics extracted from certain development data is more likely to generalize, or to provide labels, to those examples possessing similar patterns as the development data. In addition, among the covered examples, the LF may also be expected to perform more accurately on those examples within closer proximity to the development data, whereas having higher possibility of over-generalizing (to provide wrong labels) on the examples that lie in data subspaces further away from the development data. We visualize such trend in Figure~\ref{fig:lf_acc_quantile}.

\begin{figure}[!t]
    \centering
    \includegraphics[width=\linewidth]{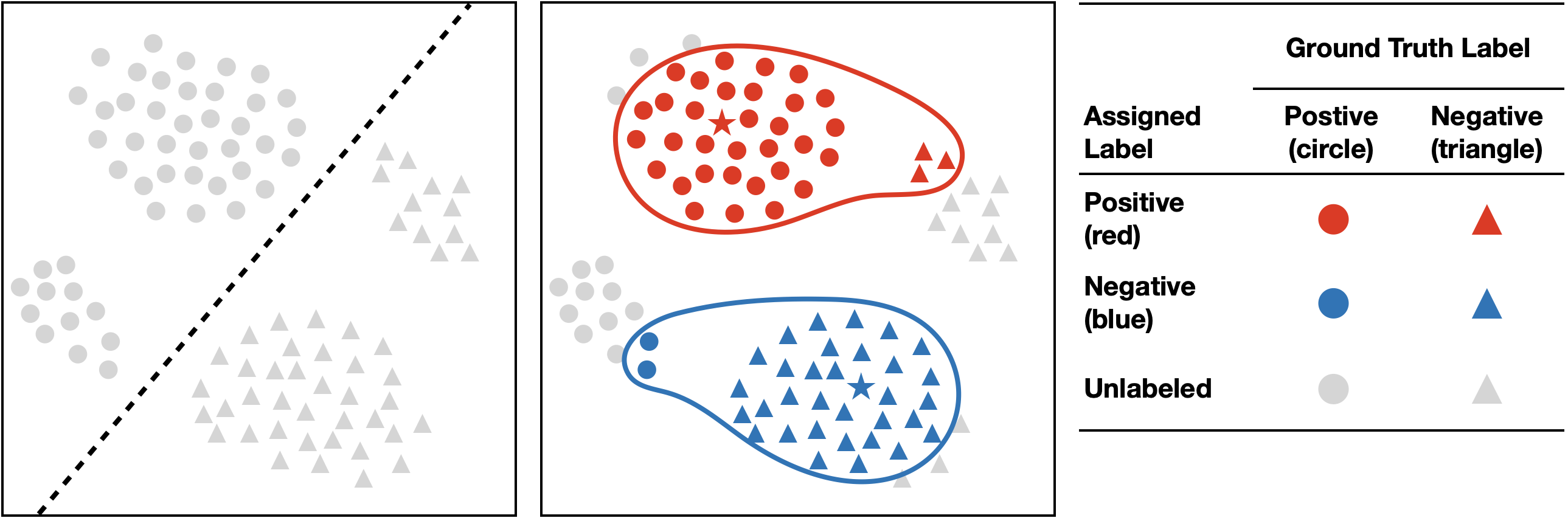}
    \vspace{-4mm}
    \caption{Left: A toy sentiment classification dataset with 4 clusters, each corresponding to product reviews from a category.
    Right: Looking at development data points (stars), users are likely to create LFs that generalize to similar examples. In addition, the LFs may be expected to be more accurate around the development data.
    Circle/Triangle corresponds to ``ground truth'' Positive/Negative label. Red/Blue corresponds to ``assigned'' Positive/Negative label and Gray is unlabeled.
    }
    \label{fig:example_lf}
    \vspace{-4mm}
\end{figure}

\begin{example}
    \label{example:data_lf_impact}
    \em
    Consider a sentiment classification task on product reviews from various categories, as illustrated in Figure~\ref{fig:example_lf}.
    First, by looking at reviews from a certain category, users would more likely create LFs that cover reviews from the same category. For instance, by looking at ``Food'' product reviews, potential LFs extracted such as ``delicious $\to$ positive'' are more likely to generalize to other ``Food'' product reviews than to ``Electronics'' product reviews. Second, an LF created from a certain category may be expected to be more accurate for reviews in the same or similar categories. For instance, by looking at reviews from the ``Movie'' category, users may find ``funny $\to$ positive'' an useful LF. However, the LF appears less accurate for reviews from ``Food'' category, since ``funny'' could indicate negative sentiment when associated with taste/food.
\end{example}

While the data samples seen in the LF development process directly impact the resultant set of LFs created and provide valuable contextual information about LFs, existing work on DP have largely considered the entire LF development process as an exogenous black-box to the DP learning pipeline~\citep{ratner2016data,ratner2017snorkel,ratner2019training,fu2020fast}. The lack of attention on the LF development process thus poses three major limitations to the current DP paradigm:
\begin{itemize}
    \item \textbf{Under-formalized LF Development Workflow:}
    The lack of formalism on the LF development process has obscured systematic study to optimize the workflow, making it less organized and more challenging for practitioners to design LFs for DP applications~\cite{varma2018snuba,galhotra2021adaptive,boecking2021interactive,zhang2022survey}.
    
    \item \textbf{Inefficient Development Data Selection:} Current LF development workflow selects development data with the most straightforward approach, uniform random sampling, which unfortunately can be time-consuming as it oftentimes requires users to inspect a considerable amount of data samples to create an informative set of LFs.
    
    \item \textbf{Dropped Data-to-LF Lineage:}
    The development context within which the LFs were developed is neglected, i.e., from which development data an LF was created, leaving behind valuable information about the LFs' expected accuracies in different data subspaces.
\end{itemize}


In this work, we make three corresponding hypotheses to tackle the limitations and improve the productivity of DP learning pipeline:
\begin{itemize}
    \item \textbf{Formalism of LF development process could allow systematic study and optimization of DP workflow.}
    By framing the LF development process formally with clear objective, we would be able to study and improve the development process, making the DP paradigm more productive.

    \item \textbf{Strategic development data selection could lead to more efficient DP pipeline.} By strategically selecting the development data to be used by the users for writing LFs, instead of random sampling, we could ideally \textit{guide} the users in efficiently creating a set of LFs that provide the most informative supervision signals for learning the subsequent models.
    
    \item \textbf{Exploitation of LF lineage to development data could enable more effective DP learning.} By tracking the LF lineages to the development data, we could ideally exploit this information to \textit{contextualize} where the LFs are expected to perform best, and accordingly model the data-conditional accuracies of the LFs for more effective denoising.
\end{itemize}

Accordingly, we then make the following technical contributions that \textit{positively} verify each of the above hypotheses:

\smallskip\noindent\textbf{Formalism of LF Development and First End-to-End System Solution:}
We formalize a broader DP paradigm, called Interactive Data Programming (IDP), that explicitly considers LF development as one central and interactive component in the entire learning workflow.
As illustrated in Figure~\ref{fig:idp}, we formulate the entire DP pipeline as an iterative human-in-the-loop cycle. In each iteration, the user creates new LFs based on the selected development data, and subsequently train the downstream model based on the LFs collected so far. The goal is to train an accurate model for the target task in as few iterations as possible.
Towards the goal, we focus on studying two core problems: (1) how to intelligently select development data so as to guide users in developing most useful LFs efficiently, and (2) how to leverage the LF lineage to development data for facilitating more effective modeling and learning from the created LFs.

With the problem formalized, we then set out to design \textit{Nemo}, the first end-to-end system for IDP which is built on top of two novel methodologies that respectively tackle the data selection and contextualized LF modeling problems.
Through extensive quantitative evaluations and a carefully conducted user study, we validate our first hypothesis where we observe that Nemo offers significant performance lift over the current prevailing DP system an average 20\%. When compared to other interactive learning schemes such as traditional active learning, Nemo as well leads to an average 34\% performance lift.

\begin{figure*}
    \centering
    \includegraphics[width=\linewidth]{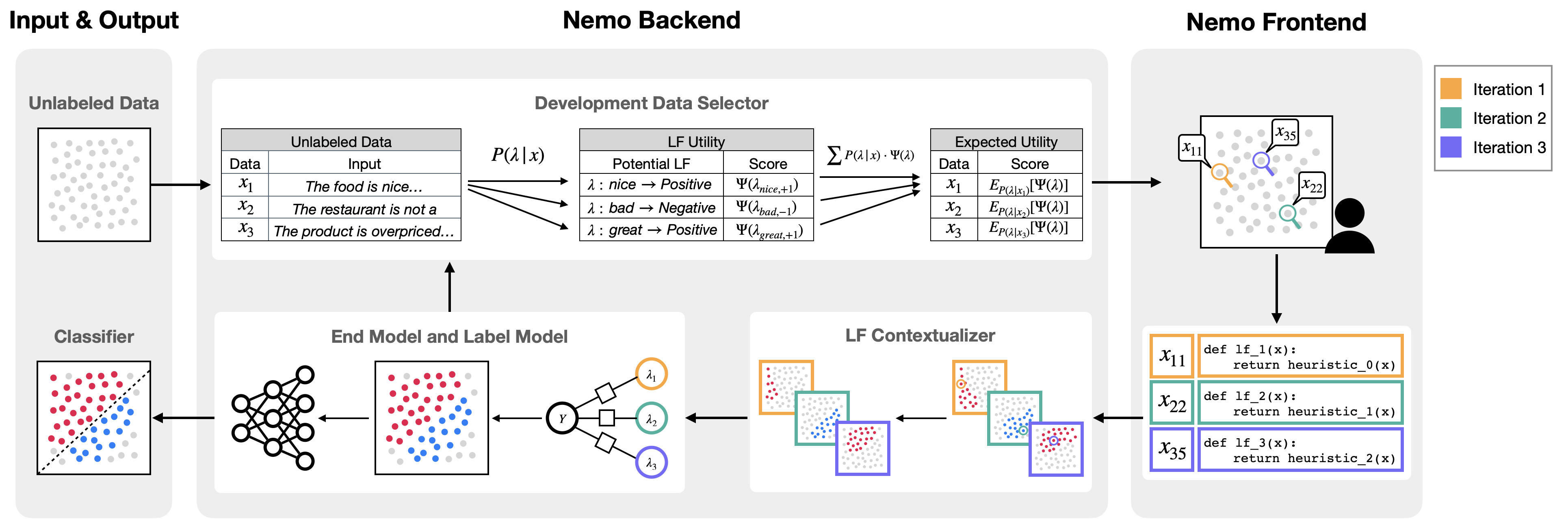}
    \vspace{-6mm}
    \caption{Nemo system overview. In each iteration, (1) the data selector intelligently picks an example from the unlabeled set based on current labeling information. (2) The user creates an LF based on the selected example. (3) The LF contextualizer refines each LF based on their development context, after which the label and end model is learned from the refined LFs.
    }
    \label{fig:idp_flow}
    \vspace{-3mm}
\end{figure*}

\smallskip \noindent \textbf{Intelligent Development Data Selection Strategy:}
In building Nemo, we propose a novel data selection strategy, \textit{Select by Expected Utility} (SEU), to efficiently guide users in developing useful LFs. The key idea within the proposed SEU approach is to first \textit{statistically measure the utilities} of the potentially generated LFs, and in turn select the examples that are expected to lead to high utility (or more informative) LFs, in which the expectation is calculated through a proposed \textit{user model} that measures the conditional probability of user returning an LF by looking a specific development data (illustrated in the development data selector in Figure~\ref{fig:idp_flow}). With SEU, we validate our second hypothesis and improve the DP performance an average 16\% compared to random sampling baseline.

\smallskip \noindent \textbf{Contextualized LF Modeling with Data Lineage:} In Nemo, we propose a model-agnostic method that exploits the LF lineage to the development data to more effectively denoise and learn from the LFs.
Particularly, based on the natural tendency for users to create LFs that are more precise around the neighborhood of the development data~\cite{Awasthi2020Learning}, we propose to refine each LF to be active only within a certain radius from their corresponding development data points (illustrated in LF contextualizer in Figure~\ref{fig:idp_flow}). With the method, we validate our third hypothesis and improve the DP performance an average 11\% compared to the standard learning pipeline without leveraging the LF development context in modeling the LFs.

\section{Preliminaries}
\label{sec:preliminaries}

We review the standard data programming (DP) setup \citep{ratner2016data,ratner2017snorkel}. 
Let each example $x \in \mathcal{X}$ be associated with a corresponding label $y \in \mathcal{Y}$, where the joint density is governed by some underlying distribution $\mathcal{D}$.
Given a set of $n$ unlabeled examples $U = \{x_i\}_{i=1}^n$, drawn from distribution $\mathcal{D}$, with their labels $\{y_i\}_{i=1}^n$ unobserved, the standard DP pipeline follows three main stages:
\begin{enumerate}
    \item \textbf{Labeling Function Development Stage:} Users encode labeling heuristics into a set of $m$ labeling functions (LFs), $\{\lambda_j\}_{j=1}^m$, where $\lambda_j: \mathcal{X} \to \mathcal{Y} \cup \{0\}$. Each LF $\lambda_j$ could either provide an example $x_i$ with a possibly noisy label $\lambda_j(x_i) \in \mathcal{Y}$ or abstain with $\lambda_j(x_i) = 0$.
    
    \item \textbf{Label Denoising/Aggregation Stage:} The provided LFs are individually applied to the unlabeled examples, generating a label matrix $L \in \mathbb{R}^{n \times m}$ where $L_{ij} = \lambda_j(x_i)$. Then, a \textit{label model} is learned to aggregate $L$, the votes on the training labels, into a set of probabilistic estimates of the ground truth labels $\{P(y_i | L)\}_{i=1}^n$.
    
    \item \textbf{End Model Learning Stage:} Finally, these probabilistic estimates could serve as probabilistic soft labels that can be used to train a final \textit{discriminative model} $f: \mathcal{X} \to \mathcal{Y}$ that could generalize and make predictions on unseen examples, with the overarching goal of minimizing the generalization error on distribution $\mathcal{D}$.
\end{enumerate}
In the LF development stage, users generally refer to some \textit{development data}, sampled from the unlabeled set $U$, to extract labeling heuristics for writing LFs.
As a result of this workflow, the data seen in the development stage has a direct influence on what LFs would be created, and on which examples these LFs may be expected to perform best (see Figure~\ref{fig:lf_acc_quantile} and Example~\ref{example:data_lf_impact}).

Despite the impact that development data have on the resultant LFs, the problems of how to strategically select these data for guiding LF development and how to potentially exploit this lineage in modeling the LFs have remained under-explored.
In fact, the entire LF development process has been given scant attention by literature to date, where the most straightforward but naive \textit{random sampling} has been the dominating approach for selecting the development data, and the modeling of LFs has been blind to their development context~\citep{ratner2016data,ratner2017snorkel,ratner2019training,fu2020fast,pmlr-v139-mazzetto21a}. Unfortunately, as discussed in previous section, this current workflow renders the entire DP pipeline less efficient and effective.

\section{Interactive Data Programming}
To study the impact of LF development process in the DP paradigm, we formalize a broader DP framework that considers LF development as one central, iterative process within the entire DP learning pipeline. We term this new formalism Interactive Data Programming (IDP), as illustrated in Figure~\ref{fig:idp}. We highlight two novelties in IDP compared to the standard DP paradigm:
\begin{itemize}
    \item First, we formalize the LF development stage as a two-step process where (1) a subset of development data would first be selected from the unlabeled set, and (2) the users then develop LFs based on these selected data. While this is an established workflow in practice, this process has not been carefully formalized and studied in the literature.
    \item Second, we consider the LF development stage and the subsequent label/end model learning stage as interleaved steps in an interactive cycle, rather than a sequential and independent procedures. In each iteration, the users develop new LFs \textit{guided} by the learning models, and the models in turn learn from the LFs created where the models are \textit{aware} of the context within which the LFs are developed.
\end{itemize}

The overarching goal of IDP is to achieve highest end model predictive performance, evaluated with a held-out \textit{test dataset}, in as few interactive iterations as possible. This entails that the users could efficiently design useful weak supervision sources, the LFs, and effectively train an accurate predictive model from the LFs.
Specifically, we study two main problems towards the goal:
\begin{itemize}
    \item First, we consider the problem of how to intelligently \textit{select} examples to \textit{efficiently} guide users in writing informative LFs from which an accurate predicitve model can be learned.
    \item Second, we consider the problem of how to exploit the \textit{LF development context} to \textit{effectively} model and learn from a given set of LFs.
\end{itemize}


\noindent \textbf{Connection to Active Learning.} We note that the first problem studied in IDP is analogous to active learning~\cite{settles2009active}, in the sense that the goal is to iteratively select data points from an unlabeled dataset and solicit supervision feedbacks from the user to train an accurate downstream model, with as few queries to the user as possible. However, unlike active learning where in each iteration the user provides supervision in terms of a \textit{single label annotation} to the selected data point, the supervision form provided in IDP is at a higher functional level (i.e., LFs) that noisily \textit{annotates multiple data points at a time}.
We emphasize that the fundamental difference between label annotations and LFs leads to a unique set of challenges in designing the data selection algorithm for IDP.
In addition, the inherent noise comes with the LFs in IDP creates a unique second problem of the need to contextualize the expected accuracy of an LF on different data subspaces, which marks another major difference to active learning.

\noindent \textbf{A Subsuming Framework.}
IDP is an encompassing framework that subsumes many existing usages of DP in practice~\citep{ratner2016data,ratner2017snorkel,cohen-wang2019interactive,Awasthi2020Learning}. For example, the current widely adopted DP workflow corresponds the vanilla instantiation of IDP that selects development data with random sampling and models the LFs without considering their development context~\citep{ratner2017snorkel}; the rule-exemplar learning approach \citep{Awasthi2020Learning} is another instance under IDP that supports half of the IDP loop where it models the LFs with their development context using the ImplyLoss model \citep{Awasthi2020Learning}, but does not consider strategically selecting the development data.

\noindent \textbf{Setup.}
Formally, given an unlabaled dataset $U = \{x_i\}_{i=1}^n$, the proposed IDP framework proceeds iteratively by the following steps:

\begin{enumerate}
    \item \textbf{Development Data Selection Stage:} In the $t$-th iteration, a subset of examples $S_t \subset U$ are strategically selected from the unlabeled set and shown to the user to guide development of LFs that are most informative to the models.
    
    \item \textbf{Labeling Function Development Stage:} Based on $S_t$, the user writes a set of $k$ labeling functions $\Lambda_t = \{\lambda_{t1}, \ldots, \lambda_{tk}\}$ which extracts meaningful labeling heuristics encoded in the examples. The lineage of these LFs to the development data $S_t$ is tracked and represented as a tuple $(\Lambda_t, S_t)$.
    
    \item \textbf{Label/End Model Learning Stage:} Given the set of LFs created so far with their data lineage, $\{(\Lambda_{1}, S_1), \ldots, (\Lambda_{t}, S_t)\}$, the label and end models are learned from the LFs as in the standard DP pipeline, but with additional access to the LFs' development context.
    Finally, the current model information would be passed back to the data selection stage to start the next cycle; or the iteration stops and the end model is output for the learning task of interest.
\end{enumerate}

\noindent \textbf{Paper Scope.} In the remainder of this paper, we focus on binary classification tasks where $\mathcal{Y} = \{-1, 1\}$ for ease of exposition. In addition, we focus on an atomic IDP setting where in each iteration, a single example is selected as the development data ($|S_t| = 1$), and the user in return provides a single LF ($|\Lambda_t| = 1$).
Despite its seemingly simplification, we emphasize that after multiple interactive iterations, the user would have seen multiple development data points, and would have developed multiple LFs out from the data points. As a result, we can see how this atomic setup effectively builds up to the general IDP setup where the user may develop multiple LFs from multiple examples in each iteration. A subtle difference lies in that, in the atomic one-example to one-LF setup, the user looks at the data points and develops LFs \textit{sequentially} rather than \textit{in batch}. This sequential nature allows most efficient use of the user's effort as the underlying development data selection algorithm can adjust dynamically \textit{given every newly developed LF}, avoiding the user spending extra effort in designing redundant LFs.
While we focus on the atomic IDP setting in the paper, we provide discussions on how our proposed solutions can be generalized to support the general IDP setup.


Finally, we note the development context of $\Lambda_t$ includes all previous sequence of development data the user has seen $(S_1, \ldots, S_t)$. In this work, we only consider the context window to include the data the user is currently looking at (i.e., $S_t$), and leave the incorporation of longer weighted context-sequence as a future direction.

\section{Nemo Architecture}
\label{sec:system}
We present Nemo, the first end-to-end system designed to support the full IDP loop by tackling the two core IDP problems.
We provide Nemo system overview in Figure~\ref{fig:idp_flow}. Nemo consists of an user-facing frontend where users develop LFs based on selected development data and a suite of backend engines where the system computes the best example to select in each iteration and learns from the created LFs along with their development context. At the frontend, Nemo provides convenient user interface for users to easily create LFs based on selected development data (Section~\ref{system:frontend}). In the system backend, the \textit{Development Data Selector} strategically selects examples from the large unlabeled set to guide users in creating informative LFs (Section~\ref{system:query_engine}). In addition, the \textit{Labeling Function Contextualizer} exploits the LF lineage to development data to facilitate more effective modeling and denoising of the LFs (Section~\ref{system:learning_engine}).

Nemo workflow begins by initially taking as input an unlabeled dataset. Then, Nemo proceeds in an interactive loop where each iteration follows the three main IDP stages as illustrated in Figure~\ref{fig:idp_flow}. We provide the system pseudo-code in Appendix A.

\noindent \textbf{System Configuration and Inputs.}
Nemo focuses on one most widely adopted type of LFs, the \textit{primitive-based LFs}.
Formally, we consider primitive-based LFs to be any function $\lambda: \mathcal{X} \to \mathcal{Y}$ that can be expressed by:
\begin{align}
    \lambda_{z,y}(x): \texttt{return $y$ if $x$ contains $z$ else abstain} \nonumber
\end{align}
where $z \in \mathcal{Z}$ is some domain-specific primitive and $y \in \mathcal{Y}$ is a target label. Such functional type (or family) of LFs has been widely considered in the literature \citep{varma2017inferring,Awasthi2020Learning,boecking2021interactive,zhang2021wrench}, and can flexibly capture arbitrary input pattern by defining the primitive domain $\mathcal{Z}$ accordingly.
One representative instantiation in text domain is the keyword-based LFs, where $\mathcal{Z}$ is a set of keywords, e.g., n-grams.
We note that the primitive-based LF form absorbs any uni-polar LF~\cite{ratner2019training} that can be expressed as $\lambda: \mathcal{X} \to \{y, 0\}$, where $y \in \mathcal{Y}$. This is because we can effectively express any LF of form $\lambda: \mathcal{X} \to \{y, 0\}$ as ``$\lambda'_{z_{\lambda},y}(x): \texttt{return $y$ if $x$ contains $z_{\lambda}$ else abstain}$'', where the \textit{contain} operator can be formally expressed as $z_{\lambda} = \mathbbm{1}_{\lambda(x)=y}$. The key idea is that the primitive domain can be arbitrary defined by the user to contain any black-box transformation on the data. Finally, we note that uni-polar LFs, that map an input to a single class or abstain, are arguably the most common type of LFs used in practice~\cite{ratner2019training}.

Before the interactive loop starts, we ask the user to configure Nemo by specifying : (1) the data domain $\mathcal{X}$, (2) the label space $\mathcal{Y}$, and (3) the primitive domain $\mathcal{Z}$.
These specifications allow Nemo to configure the user interface accordingly and later help users more conveniently develop LFs, as we shall illustrate shortly in Figure~\ref{fig:user_interface}.

\begin{example}
\em
Consider the running example of sentiment classification on product reviews. The data domain $\mathcal{X}$ is the text inputs. The label space is $\mathcal{Y} = \{\text{positive}, \text{negative}\}$. The primitive domain $\mathcal{Z}$ could be specified as the set of all uni-grams contained in the unlabeled set, which can be automatically inferred by Nemo given $U$.
\end{example}


\subsection{User Interface: Writing Labeling Functions with Development Data}
\label{system:frontend}
We describe how we design Nemo user interface to allow users conveniently create LFs guided by selected development data.

\noindent \textbf{Creating LFs with Development Data.}
In DP applications, we observe a common practice that users would go through when developing LFs based on development data. Specifically, by looking at a data sample $x$, users generally follow three principal steps to create an LF:
\begin{enumerate}
    \item Determine the corresponding label $y$ of the example $x$.
    \item Look for a primitive $z$ within $x$ that is indicative of the label $y$, which is expected to generalize well to other examples.
    \item Create and return the labeling function $\lambda_{z,y}$. 
\end{enumerate}
We note that such procedure has been widely adopted in practice and in previous literature~\citep{ratner2017snorkel,cohen-wang2019interactive,Awasthi2020Learning}. 
The Nemo user interface is designed to support this workflow where users could easily create LFs from development data through a few mouse-clicks. We demonstrate the Nemo user interface using the running example on the sentiment classification task in Example~\ref{example:ui}.

\begin{figure}
    \centering
    \includegraphics[width=\linewidth]{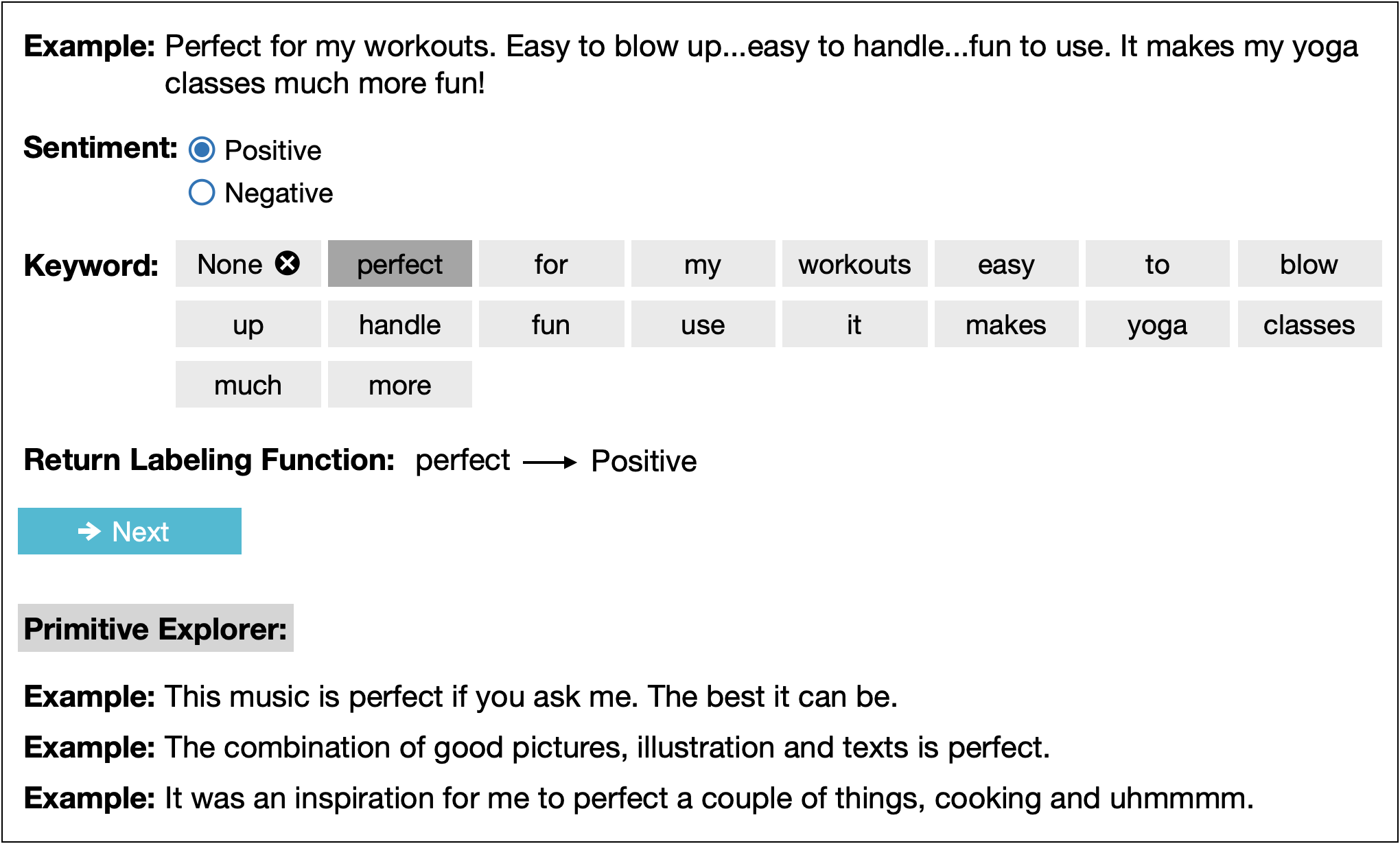}
    \caption{Nemo user interface. Users can easily create an LF from a development data by selecting a target label and a corresponding primitive.
    Note that in the example, the target labels---positive and negative---are configured according to the user provided label domain $\mathcal{Y}$. The shown candidate primitives are selected according to provided primitive domain $\mathcal{Z}$, in this case, the uni-grams in the dataset.
    }
    \label{fig:user_interface}
    \vspace{-4mm}
\end{figure}

\begin{example}
\label{example:ui}
\em
Figure~\ref{fig:user_interface} shows the Nemo user interface. In each iteration, the user will be shown a development data selected from the unlabeled set. In this case, a product review: ``Perfect for my workouts...''. To extract useful heuristic as LF from the example, the user would first determine that the review corresponds to positive sentiment. Then, the user looks for a keyword primitive in the review that supports positive sentiment, e.g., the word ``perfect''. Finally, by simply clicking on the corresponding sentiment and keyword, Nemo automatically creates an LF $\lambda_{\text{perfect, positive}}$.
\end{example}

\subsection{Development Data Selector: Guiding Informative Labeling Function Creation}
\label{system:query_engine}
The Development Data Selector is the core engine in Nemo that tackles the problem of how to guide efficient LF development through intelligently selected development data. 

\noindent \textbf{Random Selection Baseline.}
One straightforward selection approach is to \textit{randomly} sample the development data from the unlabeled set in each iteration.
In fact, as the development data selection problem has not been carefully considered in literature to date, the random selection baseline has been the prevailing approach adopted in most existing DP applications. However, a major drawback of the random approach is that it completely ignores the information provided by the current set of LFs on hand, potentially leading to the development of redundant LFs that provide little extra label information.

\begin{figure}[!t]
    \centering
    \includegraphics[width=\linewidth]{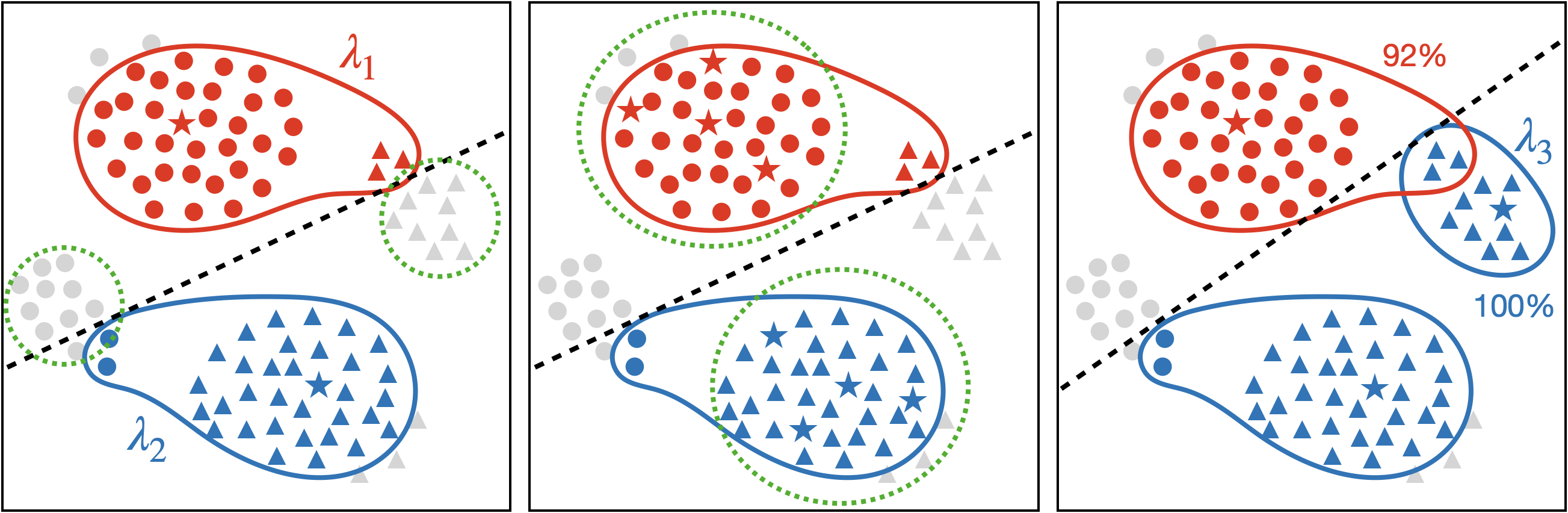}
    \caption{Left: Suppose we already have two LFs on hand. Middle: We see that random sampling may fail to solicit informative LFs from the user, where it has higher probability of selecting data points within the large, labeled clusters (the star points).
    Right: SEU selects the data point in the smaller unlabeled cluster as it has higher probability of leading to LFs that provide new, complementing label information.}
    \label{fig:example_selection}
    \vspace{-4mm}
\end{figure}

\begin{example}
\label{example:random_selection}
\em
In Figure~\ref{fig:example_selection}, suppose that the user already developed two LFs, $\lambda_1$ and $\lambda_2$, for an initially unlabeled dataset, and we are already confident about the labels in two major clusters. Ideally, we would like the user to write LFs that provide supervision over the examples that have yet received label annotations. However, by random sampling, the selected development data has high probability of being an example within the two major data clusters, given the dominating probability mass of the two major clusters. In turn, the user will likely to create an LF that annotates the examples within the two clusters, providing limited extra supervision information and starving the examples in the other two smaller clusters which as well play important roles in the classifier's decision boundary.
\end{example}

Ideally, in each iteration, we seek an LF that could best complement the current set of LFs, such that newly acquired LF could, for example, help provide coverage over an originally unlabeled data subspace, or help resolve the conflicts between current LFs. Thus, we call for an intelligent strategy that could more adaptively guide users in creating \textit{useful} LFs according to the information on hand. 

\noindent \textbf{Select by Expected Utility.}
In designing the Nemo Development Data Selector, we propose a novel selection strategy, \textit{Select by Expected Utility} (SEU), which adaptively picks the development data that \textit{in expectation}, could guide the user in developing the \textit{most informative} LF to complement the current collected label information.
Formally, in each iteration $t$, SEU selects the development data by:
\begin{align}
\label{eq:qeu}
    x^* = \argmax_{x \in U} \E_{P(\lambda|x)}[\Psi_{t}(\lambda)],
\end{align}
where $P(\lambda|x)$ is the estimated conditional probability that an user would create LF $\lambda$ given the development data being $x$, and $\Psi_t(\cdot)$ is an utility function measuring the informativeness of an LF $\lambda$ based on the supervision provided by existing set of LFs $\Lambda_t$.
SEU is shaped by two key properties in LF development:
\begin{itemize}
    \item Development Tendency: By looking at different development data, the user is likely to create different LFs.
    \item Varying LF Utilities: Different resultant LFs provide different levels of useful supervision information.
\end{itemize}
Let $\mathcal{F} = \{\lambda_{z,y} | z \in \mathcal{Z}, y \in \mathcal{Y}\}$ be the LF family where all possible LFs lie. Given a development example $x$, SEU first leverages $P(\lambda|x)$, which we call the \textit{user model}, to estimate the probability for each LF $\lambda \in \mathcal{F}$ to be created by the user. Then, SEU evaluates the informativeness of each LF $\lambda \in \mathcal{F}$ by the utility function $\Psi(\cdot)$. Finally, an example $x$'s ability to lead to an informative LF $\lambda$ can be summarized by the expected value of $\Psi(\lambda)$ taken with respect to $P(\lambda|x)$. We refer the readers to the Development Data Selector in Figure~\ref{fig:idp_flow} for an illustration of the SEU selection strategy.
By design, SEU achieves the goal of \textit{selecting an example that is expected to lead to useful LFs} by capturing and leveraging the above two key properties in LF development via the user model and utility function respectively.
We describe below how we design the user model $P(\lambda|x)$ and the utility function $\Psi(\cdot)$.

\noindent \textbf{User Model.}
The goal of user model $P(\lambda|x)$ is to provide probability estimate for the user returning an LF $\lambda$ given a development example $x$.
We model this conditional probability for any LF $\lambda_{z,y} \in \mathcal{F}$ by:
\begin{align}
    \label{eq:user_model}
    P(\lambda_{z,y} | x) = 
            \begin{cases}
                P(y) \cdot \frac{acc(\lambda_{z,y})}{\sum_{\lambda \in \{\lambda_{z,y}|\text{$z$ in $x$}\}}{acc(\lambda)}}, & \text{if $z$ contained in $x$} \\
                0, & \text{otherwise}\\
            \end{cases}
    \end{align}
where $acc(\lambda) = \frac{\sum_{\lambda(x_i) \neq 0} \mathbbm{1}_{\{{\lambda(x_i)} = \hat{y}_i\}}}{\sum_{\lambda(x_i) \neq 0}1} $ denotes the approximated accuracy of $\lambda$.
The user model closely reflects the procedure of how users develop LFs in practice.
Recall from Section~\ref{system:frontend}, when given a data sample $x$, the user would (1) determine the corresponding label $y$ for $x$, and (2) select a $y$-indicative primitive $z$ from the candidate primitives contained in $x$ to create the LF $\lambda_{z,y}$, which is expected to generalize well to other data points. 
By mirroring the procedure in parallel, the user model leverages chain rule to decompose $P(\lambda_{z,y}|x)$ into (1) the probability of $y$ being the underlying label, and (2) the probability that $z$ would be picked by the user from all candidate primitives $\{z \in \mathcal{Z}| z \text{ contained in } x\}$. In the user model, we utilize the label prior $P(y)$ to model the probability of $y$ being the ground truth label for an example $x$. Then, we model the probability that $z$ being picked to be proportional to how strongly $z$ is indicative of the target label $y$, captured by the accuracy of $\lambda_{z,y}$. In the absence of the ground truth labels, we use the current label predictions from the discriminative model $\hat{y} = f(x)$ to approximately compute the true accuracies of the LFs.
Finally, we note that the user model simply assigns zero probability to LFs that operate on primitives not residing in $x$, since these primitives would not be selected by the user from $x$.

\noindent \textbf{LF Utility Function.}
The goal of LF utility function $\Psi: \mathcal{F} \to \mathbb{R}$ is to measure the informativeness of an LF, given the current collected supervision information. We design the LF utility function to be:
\begin{align}
    \label{eq:lf_utility}
    & \Psi_t(\lambda) = \sum_{i \in C}{\psi_t^{\mathrm{uncertainty}}(x_i)} \cdot (\lambda(x_i) \hat{y}_i),
\end{align}
where $C = \{i | x_i \in U, \lambda(x_i) \neq 0\}$ is the set of indices of those examples covered by $\lambda$, and $\psi_t^{\mathrm{uncertainty}}(x_i) = -\sum_{y_i \in \mathcal{Y}} P(y_i|\Lambda_t) \cdot \log P(y_i|\Lambda_t)$ is the current \textit{label model uncertainty} on an example $x_i$ based on existing LFs $\Lambda_t$.
The utility function $\Psi(\cdot)$ gives higher scores to LFs that provide \textit{accurate} label annotations to the examples with \textit{high label uncertainty}. 
Specifically, the utility score of an LF $\lambda$ is the sum over the label uncertainty scores $\psi_t^{\mathrm{uncertainty}}(x_i)$ of the examples to which $\lambda$ provides labels, weighted by whether the provided label $\lambda(x_i)$ is correct or not, i.e., $\lambda(x_i)\hat{y}_i \in \{-1, 1\}$, where $\hat{y}_i$ is an approximate to the ground truth label.
In DP, a high label model uncertainty score $\psi_t^{\mathrm{uncertainty}}(x_i)$ corresponds to either an example that has not been covered by any LFs, or an example on which the LFs disagree the most. New label information on these uncertain data points provides informative supervision signals that reduce the label model's uncertainty over the entire dataset, complementing the existing LFs. \textit{Correct} labeling to these uncertain data points allows us to obtain a more holistic view on the entire data space or help reduce the noise within the labeled data, both leading to \textit{positive} influences to the model performances. On the other hand, \textit{incorrect} labeling on these uncertain data points introduces influential noise into the dataset, resulting in \textit{negative} impact that is likely to undermine the subsequent model performances. As a result, we design $\Psi(\cdot)$ to take into account of both the informativeness and the accuracy of the supervision an LF provides for evaluating its usefulness to the DP learning pipeline. We provide the full SEU pseudo-code in Appendix A.

\begin{example}
\label{example:SEU}
\em
In Figure~\ref{fig:example_selection}, unlike random sampling, SEU aims to select development data based on their probability of leading to useful LFs. In this case, SEU would assign higher scores to data points in the smaller, unlabeled clusters (green-dash circled area) as they have higher probability of leading to LFs that provides new supervision signals complementing current labeling information.
\end{example}

\subsection{Labeling Function Contextualizer: Modeling LFs with Development Context}
\label{system:learning_engine}
The LF contextualizer is the main component in Nemo that tackles the problem of exploiting LF development context for more effective denoising and learning from the LFs.

\noindent \textbf{Standard Learning Pipeline.}
In the DP learning pipeline without LF contextualizer, the standard procedure to learn from a given set of $m$ LFs $\{\lambda_1, \ldots, \lambda_m\}$ is to first apply each LF to the examples $\{x_i\}_{i=1}^n$ to obtain a label matrix $L$ where $L_{ij} = \lambda_{j}(x_i)$. Then, from the label matrix $L$, a label model is learned to estimate the accuracies of these LFs~\citep{ratner2016data,ratner2017snorkel,ratner2019training,fu2020fast}. The estimated accuracies are used as corresponding weighting terms in aggregating the votes provided by each LF to produce the probabilistic soft labels $\{P(y_i | L)\}_{i=1}^n$. The more accurate an LF is, the larger the weight its vote receives in the aggregation process.
Notably, in most of the prevailing label model approaches~\citep{ratner2016data,ratner2017snorkel,ratner2019training}, each LF is assumed to be uniformly accurate over all the data points it covers, i.e., each LF is modeled to have the same accuracy across the entire data space. Nonetheless, this modeling assumption is often violated in practice. Specifically, when an LF is developed by an user looking at specific development data, we observe that the LF is likely to be more accurate on the examples within closer proximity to the development data while having lower accuracy on examples that lie further away.

\begin{figure}[!t]
    \centering
    \includegraphics[width=\linewidth]{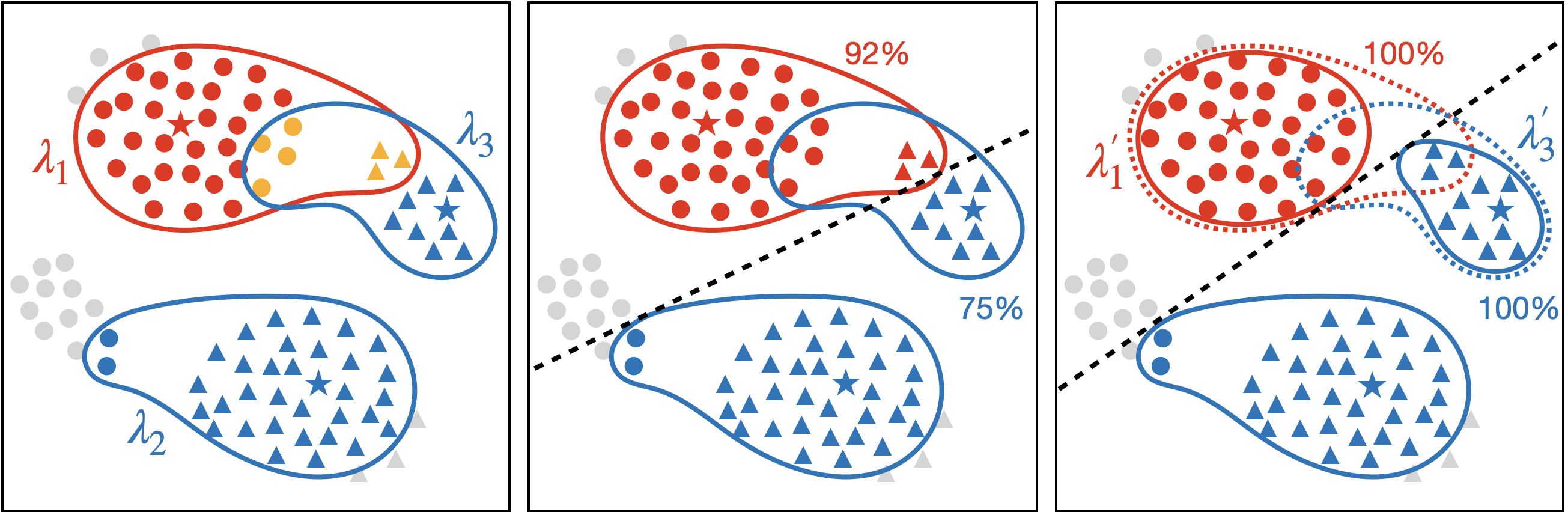}
    \caption{Left: $\lambda_1$ and $\lambda_3$ have conflicts on the yellow-colored points. Middle: In the ideal case where we could perfectly estimate the source accuracies, standard learning pipeline still fails to resolve the conflicts where it assigns wrong labels to the triangular points. Right: Contextualized learning pipeline refines the LFs and resolve the conflicts perfectly.}
    \label{fig:example_contextualize}
    \vspace{-4mm}
\end{figure}

\begin{example}
\label{example:standard_pipeline}
\em 
Continuing from Example~\ref{example:SEU}, consider a case that the newly returned LF, $\lambda_3$ in Figure~\ref{fig:example_contextualize}, is not perfectly accurate, and there are conflicts (yellow-colored) between $\lambda_1$ and $\lambda_3$ . Even if we could perfectly estimate the source accuracy of both $\lambda_1$ and $\lambda_3$, we would still assign wrong labels to the triangular points, since $\lambda_1$ would receive higher weight over $\lambda_3$ for $\lambda_1$ having a higher overall accuracy.
\end{example}

\noindent \textbf{Contextualized Learning Pipeline.}
Given the observation that LFs tend to provide noisier labels on examples that lie further away from their development data, we would ideally exploit this lineage to more effectively denoise and learn from the LFs.
To this end, we propose a contextualized learning pipeline in Nemo where we leverage the LF contextualizer to refine and denoise the LFs---according to their development context---before feeding them as input weak supervision sources to learning the subsequent label and end discriminative model, as shown in bottom pipeline in Figure~\ref{fig:idp_flow}.
In the Nemo learning pipeline, the LF contextualizer refines each LF by restricting it to be \textit{active} only on examples within a certain proximity of its corresponding development data point, dropping the labels assigned to examples that are further away as these labels are prone to be noisier.
Formally, let $x_{\lambda}$ denote the development data point from which the LF $\lambda$ is created. Given a set of $m$ LFs and their corresponding development data points $\{(\lambda_1, x_{\lambda_1}), \ldots, (\lambda_m, x_{\lambda_m})\}$, the LF contextualizer refines each LF $\lambda_j \in \{\lambda_j\}_{j=1}^m$ into $\lambda^{'}_{j}$ by:
\begin{align}
    \label{eq:refine}
    {\lambda^{'}_j}(x)  = \begin{cases}
                \lambda_j(x), & \text{if $dist(x, x_{\lambda_j}) \leq r_j$ }\\
                0 \text{ (abstain)}, & \text{otherwise}\\
            \end{cases}
\end{align}
where $dist(\cdot, \cdot): \mathcal{X} \times \mathcal{X} \to \mathbb{R}$ measures the distance  (dissimilarity) between two input examples, and $r_j$ is a given thresholding value for the refinement radius.
By discarding the assigned labels to examples where the LFs are expected to perform more poorly, we may reduce the noise in the generated label matrix $L$, and in turn produce more accurate probabilistic soft labels $\{P(y_i|L)\}_{i=1}^n$.
We provide the pseudo-code for contextualized learning in Appendix A.
In practice, the distance measurement can be selected based on feature domain. For instance, in text domain, $dist(\cdot, \cdot)$ can be the cosine distance or the euclidean distance. For the refinement radius, we set $r_j$ to be the $p$-th percentile value of $\{dist(x_{\lambda_j}, x_i)\}_{i=1}^n$, i.e., the set of all distances from each example $x_i \in U$ to the development data point $x_{\lambda_j}$, where $p$ is a system hyperparameter that can be selected based on the validation accuracy of the resultant estimated soft labels.
We note that the Nemo contextualized learning pipeline is compatible with any label modeling approach, where the LF contextualizer is essentially a pre-processing step on the weak supervision sources and the final probabilistic soft labels can be learned using any user-specified label model. This design allows Nemo to flexibly benefits from the advance in weak supervision modeling approaches, where the direction has received increasing research attention. 

\begin{example}
\label{example:contextualizer}
\em 
In Figure~\ref{fig:example_contextualize}, if we leverage the context that $\lambda_1$ and $\lambda_3$ were created from the red/blue-start points respectively, contextualized learning pipeline would be able to refine the LFs, and hopefully resolve the conflicts perfectly.
\end{example}


\vspace{-4mm}

\section{Evaluation}
\label{sec:evaluation}
We evaluate the end-to-end performance of Nemo, and perform a suite of ablation studies on (1) the proposed SEU selection strategy and (2) the contextualized learning approach. We seek to validate the following claims in response to the key hypotheses made:
\begin{itemize}
    \item \textbf{Nemo makes the end-to-end DP learning paradigm more efficient and productive.} 
    In Section~\ref{sec:exp_end_to_end}, we see that Nemo much improves over the current DP workflow by an average 20\%, validating that IDP formalism could enable optimization on DP learning paradigm.
    \item \textbf{The proposed selection strategy SEU improves the efficiency of LF development over existing selection methods.}
    In Section~\ref{sec:eval_query}, we see that SEU offers significant performance lift over the random sampling baseline by an average 16\%, validating that intelligent selection method can indeed improve the efficiency of DP pipeline and the effectiveness of SEU for this selection problem.
    \item \textbf{The proposed contextualized learning pipeline improves the learning performance over the standard learning pipeline.}
    In Section~\ref{sec:exp_learning}, we see that the contextualized learning pipeline leads to a considerable lift over standard learning pipeline by an average 11\%, validating the importance of exploiting LF development context in learning and the effectiveness of the proposed approach.
\end{itemize}

\subsection{Evaluation Setup}
\noindent \textbf{Datasets.}
We conduct experiments across five textual datasets and one image dataset, spanning three different tasks: sentiment classification, spam classification, and visual relation classification.
For sentiment classification, we include Amazon Review~\citep{he2016ups}, Yelp Review~\citep{zhang2015character}, and IMDB Review~\citep{maas2011learning}.
For spam classification, we include Youtube~\citep{youtube} and SMS~\citep{sms}.
For visual relation classification, we use the Visual Genome dataset~\citep{krishna2017visual}.
For each dataset, we randomly partition the data into 80\% training set, 10\% validation set, and 10\% test set, following the common convention \citep{zhang2021wrench}. We provide the dataset statistics in Table~\ref{table:dataset}.
Note that we only use a subset of the Visual Genome dataset that contains examples with the visual relationships of interest. Specifically, we formulate a binary classification task where the model is to classify whether an image contains the visual relationship of ``carrying'' or ``riding''.

\noindent \textbf{Evaluation Protocol.}
For performance comparisons, similar to active learning setting, we plot out the \textit{learning curve} of end model generalization performance on the test set over the number of iterations. A more efficient method could achieve higher performance within fewer number of iterations.
We measure the generalization performance using \textit{Accuracy Score} for all datasets except for SMS, in which we use \textit{F1-score} since the dataset is highly imbalanced.
For ease of comparisons, we summarize each learning curve by calculating the average performance on the learning curve, which essentially corresponds to its \textit{area under curve}. Formally, given a learning curve represented by a set of points $\mathcal{C} = \{(x_0, y_0), \ldots, (x_n, y_n)\}$ where $x_{i-1} < x_i$ and each $(x_i, y_i)$ corresponds to the model performance $y_i$ after $x_i$ iterations, we summarize the performance of the curve by: $\frac{1}{n} \sum_{i=1}^n y_i$. In the experiments, we perform a total of 50 iterations and evaluate the model performance every 5 iterations.
We include the evaluation plots in Appendix~B.
All reported results are the averaged over 5 runs with different random initializations.

\noindent \textbf{Simulated User.}
In the experiments, apart from the user study conducted in Section~\ref{sec:user_study}, we leverage simulated user to allow more extensive evaluations of the methods.
Similar to \citep{boecking2021interactive}, we utilize ground truth labels to simulate real user feedbacks.
Specifically, when an example $x_i$ is selected and $x_i$ contains a set of primitives $C = \{z \in \mathcal{Z} | z \text{ contained in } x_i  \}$, we first build a set of candidate LFs $\Lambda = \{\lambda_{z, y_i} | z \in C\}$.
Then, to resemble human expertise, the candidate set $\Lambda$ is refined by filtering out LFs whose accuracy is lower than some threshold $t$.
Finally, one LF is sampled from the refined set of candidate LFs and returned~\footnote{When available, we leverage external lexicon (e.g., opinion lexicon for sentiment~\cite{Hu2004MiningAS}) to further simulate real user responses. More details in Appendix C.}. We set $t=0.5$ in the experiments if not otherwise mentioned.


\noindent \textbf{Implementation Details.}
In the experiments, we featurize the input text examples with their TF-IDF representation for text datasets, and use pre-trained Resnet~\citep{he2016deep} to extract feature for image examples. We fix the end model to be logistic regression model for all methods. If not otherwise mentioned, we adopt MeTaL~\citep{ratner2019training} as the underlying label model to aggregate the weak labels. We consider the primitive domain $\mathcal{Z}$ to be the set of uni-grams in training examples for text datasets. For the image dataset, we utilize the associated object annotations as the corresponding primitives for each image. In practice, such object annotations can be obtained by leveraging any off-the-shelf object detection models.
We include more details in Appendix~C.

\begin{table}[t]
	\centering
	\caption{Dataset statistics. ``Cls.'' stands for classification task.}
	\vspace{-4mm}
        \begin{tabular*}{\linewidth}{llccc}
        		\toprule
\textbf{Task} & \textbf{Dataset}  & \textbf{\#Train} & \textbf{\#Valid}  & \textbf{\#Test} \\
\midrule
\multirow{3}{*}{Sentiment Cls.} & Amazon & 14,400 & 1,800 & 1,800\\
& Yelp &  20,000 & 2,500 & 2,500\\
& IMDB & 20,000 & 2,500 & 2,500\\\midrule
\multirow{2}{*}{Spam Cls.} & Youtube & 1,566 & 195 & 195 \\
& SMS & 4,458 & 557 & 557\\
\midrule
Visual Relation Cls. & VG & 5,084 & 635 & 635 \\

		\bottomrule
         \end{tabular*}
    \label{table:dataset}
    \vspace{-4mm}
\end{table}

\begin{table*}[!t]
\caption{Performances of Nemo and existing baselines across datasets. We see Nemo consistently outperforms the baselines, and that the proposed IDP framework offers strong performance when compared to other existing interactive schemes.}
\label{table:end_to_end}
\centering
\vspace{-4mm}
\begin{tabular*}{\linewidth}{l@{\extracolsep{\fill}}ccccccccc}
\toprule
 & \multicolumn{8}{c}{Methods}\\
\cmidrule(lr){2-10}
&  \multicolumn{1}{c}{Full IDP} & \multicolumn{1}{c}{Vanilla IDP} & \multicolumn{2}{c}{Selection-only IDP} & \multicolumn{1}{c}{CL-only IDP} & \multicolumn{4}{c}{Other Interactive Schemes} \\
\cmidrule(lr){2-2} \cmidrule(lr){3-3} \cmidrule(lr){4-5} \cmidrule(lr){6-6} \cmidrule(lr){7-10}
Dataset & Nemo  & Snorkel & Snorkel-Abs & Snorkel-Dis & ImplyLoss-L &  US & IWS-LSE & BALD & AW\\
\midrule
Amazon & \textbf{0.7674} & 0.6774 & 0.6783 & 0.6733 & 0.6822 & 0.5970 & 0.6234 & 0.6193 & 0.6951 \\
Yelp & \textbf{0.7907} & 0.6556 & 0.6664 & 0.6887 & 0.7009 & 0.6239 & 0.6415 & 0.6129 & 0.6745  \\
IMDB & \textbf{0.7958} & 0.7107 & 0.7338 & 0.7480 & 0.6766 & 0.6058 & 0.6295 & 0.5933 & 0.7247 \\
Youtube & \textbf{0.8722} & 0.8235 & 0.8541 & 0.8527 & 0.6811 & 0.7609 &  0.7904 & 0.7816 & 0.8073 \\
SMS & \textbf{0.7038} & 0.4789 & 0.6189 & 0.5485 & 0.5065 &  0.4234 & 0.6305 & 0.4536 & 0.5569 \\
VG & \textbf{0.6701} & 0.6152 & 0.6250 & 0.6384 & 0.6270 & 0.5662 & 0.5976 & 0.5703 & 0.5914 \\
\bottomrule

\end{tabular*}

\end{table*}

\begin{table*}[!t]
\caption{Performances and median user response time for different methods in the user study on Amazon dataset. Nemo significantly outperforms the baselines while taking slightly more time for users to respond compared to US and IWS-LSE.}
\label{table:user_study}
\centering
\vspace{-4mm}
\begin{tabularx}{\linewidth}{@{}l@{\extracolsep{\fill}}CCCCC@{\extracolsep{\fill}}CC@{\extracolsep{\fill}}}
\toprule
 & \multicolumn{7}{c}{Methods}\\
\cmidrule(lr){2-8}
&  \multicolumn{1}{c}{Full IDP}  & \multicolumn{1}{c}{Vanilla IDP} & \multicolumn{2}{c}{Selection-only IDP} & \multicolumn{1}{c}{CL-only IDP} & \multicolumn{2}{c}{Other Interactive Schemes} \\
\cmidrule(lr){2-2} \cmidrule(lr){3-3} \cmidrule(lr){4-5} \cmidrule(lr){6-6} \cmidrule(lr){7-8}
Metric & Nemo  & Snorkel & Snorkel-Abs & Snorkel-Dis & ImplyLoss-L &  US & IWS-LSE \\
\midrule
Performance & \textbf{0.7473} & 0.6665 & 0.6689 & 0.6600 & 0.6833 & 0.5882 & 0.5971\\
React Time (Median) & 14.42s & 16.21s & 17.95s & 13.05s & 16.21s & 12.50s & \textbf{6.73s} \\

\bottomrule
\end{tabularx}

\end{table*}

\subsection{Nemo End-to-End System Performance}
\label{sec:exp_end_to_end}
We demonstrate that Nemo outperforms existing baseline methods across various datasets through extensive quantitative experiments and an user study. We validate the importance of different Nemo system components, and demonstrate Nemo's robustness to the change of LFs' accuracy level.

\noindent \textbf{Comparisons to baseline methods.}
We include two sets of baseline methods in our experiments. First, we include existing methods that are subsumed under IDP framework:
\begin{itemize}
    \item \textit{Snorkel}~\citep{ratner2017snorkel}: Snorkel is a vanilla instantiation of IDP that selects development data by random sampling, and learns from LFs without utilizing their development context. It is the predominant approach used in practice.
    
    \item \textit{Snorkel-Abs}~\citep{cohen-wang2019interactive}: Snorkel-Abs is a selection-only IDP method that adaptively selects development data on which the current LFs abstain the most. It learns from LFs without considering their development context.
    
    \item \textit{Snorkel-Dis}~\citep{cohen-wang2019interactive}: Snorkel-Dis is as well a selection-only IDP approach that stratigically selects development data on which the current LFs disagree the most. It does not leverage LF development context in learning.
    
    \item \textit{ImplyLoss-L}~\footnote{Note that we modify the discriminative part of the ImplyLoss model to be a linear model (hence the suffix ``L'') for consistency across the methods.}~\citep{Awasthi2020Learning}: ImplyLoss-L is a contextualized-learning only IDP approach. It learns from LFs with their development context modeled through a deliberately designed model and loss function. However, it does not consider strategically selecting the development data. Thus, we couple it with the random sampling selection method.
\end{itemize}
Second, we include methods under other related interactive schemes:
\begin{itemize}

    \item \textit{Uncertainty Sampling (US)}~\citep{lewis1994sequential}: US is a classic and competitive method within active learning paradigm~\cite{settles2009active}.
    
    \item \textit{BALD}~\citep{houlsby2011bayesian,gal2017deep}: BALD is a representative bayesian active learning method that has been used as a strong active learning baseline in recent literature.
    
    \item \textit{IWS-LSE}~\citep{boecking2021interactive}: IWS-LSE is a representative method under the interactive weak supervision paradigm considered in \citep{galhotra2021adaptive,boecking2021interactive}, where
    the user is iteratively queried to provide feedback on whether a suggested labeling heuristic is useful or not.
    
    \item \textit{Active Weasul (AW)}~\citep{biegel2021active}: AW combines active learning with weak supervision by asking the user to hand-label selected data points in order to help the label model in denoising a \textit{fixed set} of LFs. Note that AW requires an initial given set of LFs. In the experiments, we use Snorkel in the first 10 iterations to generate the LF set.
\end{itemize}

In Table~\ref{table:end_to_end}, we see that Nemo consistently outperforms all the baseline methods by a significant margin, with an average 9\%  performance lift over the second-best performing methods in each dataset.
More closely, by comparing Nemo to Snorkel, we observe an average 20\% improvement, verifying the importance and benefits of the IDP framework that considers optimizing and exploiting LF development process in the DP pipeline.
While Snorkel-Abs and Snorkel-Dis improve over vanilla Snorkel by considering more adaptive data selection methods, the performance lift is much smaller in these cases. This is largely due to the heuristic design of the two approaches and their lack of further leveraging LF development context in learning.
For ImplyLoss-L, although it improves over Snorkel (in 3 out of 5 datasets) by learning with contextualized LFs, without strategically selecting the development data, its performance lift appears limited especially when compared to Nemo.
In sum, while previous methods had improved over the vanilla Snorkel baseline by either (1) designing better selection strategies or (2) learning with LF development context, none of these methods have considered both problems in an unifying interactive framework, thus rendering limited performance improvement when compared to Nemo which tackles both problems simultaneously in IDP.

From Table~\ref{table:end_to_end}, we also observe that by soliciting user supervision at functional level, i.e., LFs, IDP methods (including the vanilla Snorkel) achieve better performances over US that queries for user feedback at single label annotation level, by up to 66\% improvement. Similarly, IDP methods generally perform preferably against IWS-LSE that queries for user feedback on the usefulness of single labeling heuristic per iteration.
Notably, the real user effort spent in answering each query in different interactive schemes may not be completely reflected by the reported performances, since varying interactive schemes require different types of user inputs.
However, we believe that the results indeed demonstrate the importance and the promising potential of studying the proposed IDP framework.

\begin{figure*}[!t]
    \centering
    \includegraphics[width=\linewidth]{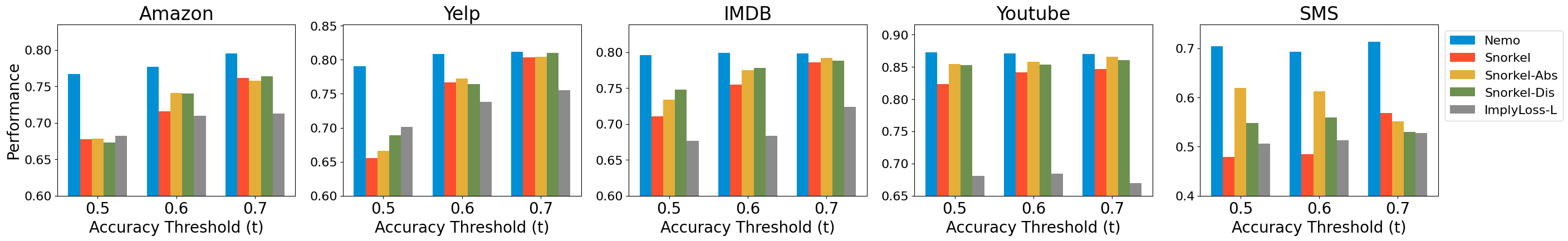}
    \vspace{-6mm}
    \caption{Performances under varying LF accuracy thresholds. Nemo demonstrates the strongest robustness when LF accuracy threshold decreases from 0.7 to 0.5.}
    \label{fig:lf_acc}
\end{figure*}

\noindent \textbf{Case study with real users.}
\label{sec:user_study}
We conduct a carefully designed user study to validate the effectiveness of Nemo with real users developing the LFs based on selected development data.
In the user study, we invited a total of 15 users with knowledge in basic machine learning as participants, including graduate students and industry engineers working in related fields. In the study, each user is asked to perform 2 randomly assigned interactive learning tasks, so that each method receives results from 5 users. We compute the result for ImplyLoss based on LFs created in the Snorkel user study, since random selection strategy is used in both approaches.
In each task, the user goes through a total of 30 interactive iterations, and we record the model performance every 3 iterations. We randomly shuffle the order an user perform on the 2 assigned tasks to avoid potential bias that could be introduced by the ordering of different methods. Before the start of each study, we gave the user a short introduction about the background of Data Programming.

We perform the study using the sentiment classification task on Amazon Review dataset which most users are familiar with. We report the user study results in Table~\ref{table:user_study}. We observe an overall similar trend to our findings with simulated users. Specifically, Nemo provides a significant performance lift, up to 13\% and 27\%, over existing DP and interactive methods respectively. This showcases that Nemo, by supporting the full IDP loop, indeed leads to an efficient and productive learning workflow in practice with real users.
In addition, we see that users generally spend a little more time in providing LFs as feedback in IDP setting than providing label annotation as response in active learning setting. The overhead lies in the extra time, which is roughly around 2 to 3 seconds per iteration, that an user needs to select a corresponding primitive from the development data in addition to determining its ground truth label.
We also note that IWS-LSE has the shortest user response time, as determining whether an LF is useful or not is generally easier than determining the label for a data point.

\begin{table}[!t]
\caption{Comparisons between Nemo performance with and without either the data selector or the LF contextualizer. We see that both components are critical to Nemo.}
\label{table:nemo_ablation}
\centering
\vspace{-4mm}
\begin{tabular*}{\linewidth}{lccc}
\toprule
 &  & \multicolumn{2}{c}{Ablated Version}\\
\cmidrule{3-4}
Dataset & Nemo & No Data Selector & No LF Contextualizer \\
\midrule
Amazon & \textbf{0.7674} & 0.7244 & 0.7384 \\
Yelp & \textbf{0.7907} & 0.7360 & 0.7219 \\
IMDB & \textbf{0.7958} & 0.7557 & 0.7932 \\
Youtube & \textbf{0.8722} & 0.8407 & 0.8628\\
SMS & \textbf{0.7038} & 0.6092 & 0.6899 \\
VG & \textbf{0.6701} & 0.6253 & 0.6542 \\
\bottomrule
\end{tabular*}
\vspace{-4mm}
\end{table}

\noindent \textbf{Ablation study on Nemo.}
We study the importance of the two core components in Nemo, i.e.,  the development data selector and the LF contextualizer.
Specifically, we evaluate Nemo's performance when either one of the components is removed. We see that in Table~\ref{table:nemo_ablation}, removing either components decreases Nemo's performance, with an average 7\% and 3\% drop when the data selector and the LF contextualizer is removed respectively.
This showcases the importance of both components in Nemo and the need to consider both the data selection and contextualized learning problems in IDP.


\noindent \textbf{Sensitivity to LF precision.}
\label{sec:lf_acc}
We evaluate the robustness of Nemo with respect to the accuracy of input LFs. Recall that when simulating real user response, the oracle simulator filters out LFs with accuracy lower than a given threshold $t$. Here, we evaluate Nemo under varying values of $t$, along with comparisons to other baseline IDP methods. From Figure~\ref{fig:lf_acc}, we first observe that as the threshold value increases, there is an overall trend of performance improvements for all methods, suggesting that, perhaps unsurprisingly, users could in general improve DP learning performance by providing more accurate LFs. Next, we see that regardless of the accuracy threshold values, Nemo consistently achieves the best performance as compared to other baselines. Finally, Nemo demonstrate stronger robustness to the threshold value change than the baseline methods. For example, we see that the performance of Snorkel drastically drops when the threshold value decreases from $0.7$ to $0.5$, whereas Nemo remains much stable across different threshold values.
The results suggest that Nemo could be more reliably deployed in practice where the accuracies of LFs developed could vary in range.

\subsection{SEU Selection Strategy Performance}
\label{sec:eval_query}

We evaluate the effectiveness of the proposed selection strategy SEU by comparing it to different baseline selection approaches, and ablating various aspects of its design. To focus on the comparisons between different selection strategies alone, we fix the learning pipeline to be the standard vanilla procedure (without the use of LF development context) in the following sets of experiments.

\begin{table}[!t]
\caption{Performances of different selection strategies when the learning pipeline is fixed to be the standard vanilla approach without using LF development context. We see that SEU consistently outperforms, by a large margin, the other baselines in all datasets considered.}
\label{table:query_methods}
\centering
\vspace{-4mm}
\begin{tabular*}{\linewidth}{l@{\extracolsep{\fill}}cccc}
\toprule
 & \multicolumn{4}{c}{Selection Strategy}\\
\cmidrule{2-5}
Dataset & SEU & Random & Abstain & Disagree\\
\midrule
Amazon & \textbf{0.7384} & 0.6774 & 0.6783 & 0.6733 \\
Yelp & \textbf{0.7219} & 0.6556 & 0.6664 & 0.6887 \\
IMDB & \textbf{0.7932} & 0.7107 & 0.7338 & 0.7480\\
Youtube & \textbf{0.8628} & 0.8235 & 0.8541 & 0.8527\\
SMS & \textbf{0.6899} & 0.4789 & 0.6189 & 0.5485 \\
VG & \textbf{0.6542} & 0.6152 & 0.6250 & 0.6384\\
\bottomrule
\end{tabular*}
\vspace{-4mm}
\end{table}

\begin{table}[!t]
\caption{Ablation study on the SEU's user model. We see that the accuracy-weighted design is critical to SEU's success.}
\label{table:user_model}
\centering
\vspace{-4mm}
\begin{tabular*}{\linewidth}{l@{\extracolsep{\fill}}cc}
\toprule
 & \multicolumn{2}{c}{User Model}\\
\cmidrule{2-3}
Dataset & SEU (\eqref{eq:user_model}) & Uniform \\
\midrule
Amazon &  \textbf{0.7384} & 0.6774\\
Yelp & \textbf{0.7219} & 0.6556\\
IMDB & \textbf{0.7932} & 0.7107\\
Youtube & \textbf{0.8628} & 0.8235\\
SMS & \textbf{0.6899} & 0.4789\\
VG & \textbf{0.6542} & 0.5592 \\
\bottomrule
\end{tabular*}
\vspace{-4mm}
\end{table}

\noindent \textbf{Comparisons to baseline selection approaches.}
We compare SEU to three baseline selection methods: 
\begin{itemize}
    \item The \textit{Random} baseline that selects randomly from the unlabeled set~\cite{ratner2017snorkel}.
    \item The \textit{Abstain} baseline that selects the data point on which the current LFs abstain the most~\cite{cohen-wang2019interactive}.
    \item The \textit{Disagree} baseline that selects the data point on which the current LFs disagree the most~\cite{cohen-wang2019interactive}.
\end{itemize}
From Table~\ref{table:query_methods}, we see that SEU consistently enjoys better performance than the other baseline selection approaches often by a large margin. 
When compared to Random, SEU provides performance lift up to 44\%, validating its capability of strategically selecting useful development data to guide efficient development of LFs.
When compared to Abstain and Disagree, SEU as well enjoys improvement up to 26\%, demonstrating its advantage over these existing straightforward strategies.

\noindent \textbf{Ablation study on user model.}
Recall that in SEU's user model (\eqref{eq:user_model}), the conditional probability $P(\lambda_{z,y} | x)$, if not zero, is modeled based on the estimated accuracy of $\lambda_{z,y}$. SEU models that an user would have higher probability of picking a primitive $z$ from $x$ to create an LF $\lambda_{z,y}$ if $\lambda_{z,y}$ has higher accuracy.
Here, we examine the importance of this accuracy-weighted design in the user model by comparing to a baseline where we modify the user model to \textit{not consider} the accuracies of LFs and to essentially assign \textit{uniform} probability to the potential LFs $\{\lambda_{z,y} | z \in \mathcal{Z}, z \text{ contained in  } x\}$:
\[
    P(\lambda_{z,y} | x) = 
            \begin{cases}
                P(y) \cdot \frac{1}{\sum_{\lambda \in \{\lambda_{z,y}|\text{$z$ in $x$}\}}{1}}, & \text{if $z$ contained in $x$} \\
                0, & \text{otherwise}\\
            \end{cases}
\]
From Table~\ref{table:user_model}, we see that it is indeed important to model $P(\lambda_{z,y} | x)$ differently based on the accuracy of $\lambda_{z,y}$, where the performance of SEU much degrades when $P(\lambda_{z,y} | x)$ is modeled uniformly for all possible LFs.

\begin{table}[!t]
\caption{Ablation study on SEU's LF utility function. We see that SEU achieves best performances when the utility function captures both the informativeness and the correctness aspects of an LF.}
\label{table:lf_utility}
\centering
\vspace{-4mm}
\begin{tabular*}{\linewidth}{l@{\extracolsep{\fill}}ccc}
\toprule
 &  \multicolumn{3}{c}{LF Utility Function}\\
 \cmidrule{2-4}
Dataset &  SEU (\eqref{eq:lf_utility}) & No Informativeness & No Correctness  \\
\midrule
Amazon & \textbf{0.7384} & 0.7369 & 0.6683\\
Yelp & \textbf{0.7219} & 0.7211 & 0.6536\\
IMDB & \textbf{0.7932} & 0.7911 & 0.7847\\
Youtube & \textbf{0.8628} & 0.8538 & 0.8552\\
SMS & \textbf{0.6899} & 0.6695 & 0.6517\\
VG & \textbf{0.6542} &  0.6486 &  0.6346 \\
\bottomrule
\end{tabular*}
\vspace{-4mm}
\end{table}

\noindent \textbf{Ablation study on LF utility function.}
Another core component in SEU is the LF utility function, which is designed to assign higher score to LFs that provide informative and accurate supervision signals. Specifically, recall that in \eqref{eq:lf_utility}, the first term $\psi_t^{\mathrm{uncertainty}}(x_i)$ and the second term $\lambda(x_i) \hat{y}_i$ capture the informativeness and the correctness of the provided label $\lambda(x_i)$ respectively.
We evaluate the importance of both aspects by comparing to two ablated versions of utility functions, in which we remove either term respectively:
\begin{itemize}
    \item No Informativeness: $\Psi_t(\lambda) = \sum_{i \in C} \lambda(x_i) \hat{y}_i$
    \item No Correctness: $\Psi_t(\lambda) = \sum_{i \in C}{\psi_t^{\mathrm{uncertainty}}(x_i)}$
\end{itemize}
In Table~\ref{table:lf_utility}, we see that removing either the informativeness or the correctness term in the utility function decreases SEU performance, suggesting the importance of considering both properties when measuring an LF's utility.

\begin{table}[!t]
\caption{Performances of different approaches to learn from LFs. We see that contextualized pipeline improves over the standard pipeline and the ImplyLoss model.}
\label{table:cl_baselines}
\centering
\vspace{-4mm}
\begin{tabular*}{\linewidth}{l@{\extracolsep{\fill}}ccc}
\toprule
 & \multicolumn{3}{c}{Learning Approach}\\
\cmidrule{2-4}
Dataset  & Contextualized & Standard & ImplyLoss  \\
\midrule
Amazon &  \textbf{0.7244} & 0.6774 & 0.6822\\
Yelp &  \textbf{0.7360} & 0.6556 & 0.7009 \\
IMDB & \textbf{0.7557} & 0.7107 & 0.6766\\
Youtube &  \textbf{0.8407} & 0.8235 & 0.6811\\
SMS &  \textbf{0.6092} & 0.4789 & 0.5065\\
VG & 0.6253 & 0.6152 & \textbf{0.6270}\\
\bottomrule
\end{tabular*}
\vspace{-4mm}
\end{table}

\subsection{Contextualized Learning Performance}
\label{sec:exp_learning}
We evaluate the effectiveness of the proposed contextualized learning pipeline, and ablate different aspects of its design choices. To focus on the comparisons between different approaches to learn from the LFs, we fix the data selection strategy to be the vanilla random selection approach in the following sets of experiments.

\noindent \textbf{Comparisons to baseline methods.}
We compare the contextualized learning pipeline to two baseline approaches:
\begin{itemize}
    \item The \textit{Standard} learning pipeline that learns from the LFs without utilizing any LF contextual information.
    \item The \textit{Implyloss} learning model that learns from the LFs and their contextual information through a specialized model and loss function.
\end{itemize}
In Table~\ref{table:cl_baselines}, we demonstrate that the proposed contextualized learning pipeline can effectively leverage the LF development context to better model and learn from the weak supervision sources. It improves over the standard learning pipeline by up to 27\%. In particular, by only refining the LFs' coverage and learn with the same underlying label model (MeTaL), we obtain larger performance improvement compared to the improvement brought by designing a more sophisticated ImplyLoss model.

\noindent \textbf{Ablation study on distance function.} Recall that in the contextualized learning pipeline, the LF contextualizer relies on a distance function to refine the LFs. We compare how different distance measurements affect the performance. In Table~\ref{table:cl_distance}, we see that Cosine distance generally brings larger performance lift than Euclidean distance. We note that regardless of the distance function used, the contextualized pipeline improves over the standard counterpart.




\begin{table}[!t]
\caption{Contextualized learning with different distance functions. Cosine distance offers larger lift than Euclidean distance while both improves over standard pipeline.}
\label{table:cl_distance}
\centering
\vspace{-4mm}
\begin{tabular*}{\linewidth}{l@{\extracolsep{\fill}}ccc}
\toprule
 & \multicolumn{2}{c}{Contextualized} & \\
\cmidrule{2-3}
Dataset  & Cosine Distance & Euclidean Distance  &  Standard \\
\midrule
Amazon & \textbf{0.7244} & 0.6913 & 0.6774 \\
Yelp & \textbf{0.7360} & 0.6991 & 0.6556\\
IMDB & \textbf{0.7557} & 0.7200 & 0.7107\\
Youtube & \textbf{0.8407} & 0.8181 & 0.8235\\
SMS & 0.6092 & \textbf{0.6174} & 0.4789\\
VG & 0.6253 &  \textbf{0.6332} & 0.6152\\
\bottomrule
\end{tabular*}
\vspace{-4mm}
\end{table}

\section{Related Work}

Recent progress in DP has largely been made in developing advanced label models for various applications \citep{ratner2016data,ratner2019training,fu2020fast,pmlr-v139-mazzetto21a,cachay2021endtoend,bach2019snorkel,safranchik2020weakly,ren2020denoising,das2020goggles,hooper2020cut,Varma2019multi,shin2021universalizing}.
We defer readers to \cite{zhang2022survey} for a more comprehensive survey on weak supervision methods.

\noindent \textbf{Labeling Function Development.}
To reduce user effort spent in LF development, studies have mainly taken three directions: (1) \textit{automatic LF generation}, (2) \textit{interactive LF discovery}, and (3) \textit{interactively-guided LF development} which our work falls into.
Automatic LF generation methods aim to create LFs automatically. The methods generally require an initial set of labeled data, or seed LFs developed by users. Snuba~\cite{varma2018snuba} learns weak classifiers as heuristic models from a small labeled dataset; TALLOR~\cite{TALLOR} and GLaRA~\cite{glara} use an initial set of seed LFs to generate new ones by compounding multiple simpler LFs and by exploiting the semantic relationship of the seed LFs respectively; \cite{tseng2021automatic} applies program systhesis to generate task-level LFs from a set of labeled data and domain-level LFs.
Interactive LF discovery methods consider adaptively searching for useful LFs from a large candidate set, by interactively querying for user's feedback on whether some suggested LFs are useful or not. The candidate LFs are generated based on context-free grammar~\citep{galhotra2021adaptive}, n-grams~\citep{boecking2021interactive}, or pretrained language models~\cite{zhang2022prboost}. Based on the user's feedback on the usefulness (whether an LF is better than random) of some selected LFs, the systems adapt and learn to identify promising LFs from the large candidate set, which are output as the final LFs to be used in the subsequent DP pipeline. 
Unlike the above two directions that require different forms of user inputs in LF development process as compared to the existing workflow, our work takes the third direction (interactively-guided LF development), which is built to inherently support the existing workflow used in practice, i.e., users write LFs by drawing ideas from development data.
However, instead of simply selecting the development data randomly from the unlabeled set, this direction considers strategically selecting informative development data to guide the users in designing useful LFs efficiently. Within this direction, \cite{cohen-wang2019interactive} performed an initial exploration, but in a relatively ad-hoc way. Our work proposes the first formalism, IDP, for the problem that further extends the scope to exploiting the information in LF development process to better model the resultant LFs.

\noindent \textbf{Connecting Data Programming and Active Learning.}
Related to our work, prior studies have explored the connection between DP and active learning from other perspectives: (1) applying active learning to complement DP, and (2) leveraging DP techniques for active learning.
Within the first direction, \cite{Maheshwari2021SemiSupervisedDP} proposed to complement an existing set of LFs by asking users to annotate a selected subset of unlabeled data;
Similarly, \cite{biegel2021active} asks users to label selected data points that would be most informative in helping denoise and aggregate the weak supervision sources in DP;
Asterisk \cite{nashaat2020asterisk} employed an active learning process to enhance the quality, in terms of accuracy and coverage, of the weak labels initially provided by the LFs.
On the second direction, \cite{Nashaat2018HybridizationOA} applied DP to generate an initial set of weak labels to improve the efficiency of a later active learning process;
\cite{Mallinar2020IterativeDP} aimed to expand an initial set of labeled data with examples from another larger unlabeled dataset. The method first constructs neighborhood-based LFs from the seed data, and utilizes the LFs to identify relevant candidate examples from the larger unlabeled set, where the candidate examples are finally annotated by application users.

\vspace{-1mm}
\section{Discussion}
While we have primarily focused on the atomic one-example to one-LF IDP setup, in this section, we discuss how Nemo can be flexibly extended to support the general setup wherein users can leverage multiple examples to create multiple LFs in each IDP iteration.
Specifically, the extension includes (1) a system-level feature that allows users to use multiple examples when creating an LF, and (2) an algorithmic-level redesign that enables Nemo to model the probability of the user returning multiple LFs per iteration.

\noindent \textbf{Primitive-based Example Explorer.}
One limitation on the atomic IDP setup is that users might sometimes find it hard to develop a new LF by looking at a single example, as it may be difficult to judge how well an LF can generalize to other examples. As a result, we enrich the Nemo user interface with the \textit{primitive-based example explorer}.
As shown in Figure~\ref{fig:user_interface}, when shown a development data point, the user can click on the candidate primitives and utilize the example explorer to view a randomly sampled set of examples containing the selected primitive. In this way, the user is able to leverage additional data points to evaluate the quality of an LF.

\noindent \textbf{Multi-LF User Model.} We demonstrate how SEU can be generalized to accommodate the IDP setup where the user may return multiple LFs in each iteration. First, we modify the SEU utility measurement (\eqref{eq:qeu}) to the following:
\begin{align}
    x^* = \argmax_{x \in U} \E_{P(\Lambda|x)}[\sum_{\lambda \in \Lambda}\Psi_{t}(\lambda)],
\end{align}
Note that we replace $\lambda$ (a single LF) with $\Lambda$ (a set of LFs). In addition, we measure the usefulness of the returned set of LFs by summing over the individual utilities of the included LFs.
Then, we can measure the user model by $P(\Lambda | x) = \Pi_{\lambda \in \Lambda} P(\lambda | x)$, where:
\begin{align}
    \label{eq:user_model}
    P(\lambda_{z,y} | x) = 
            \begin{cases}
                P(y) \cdot acc(\lambda_{z,y}) \cdot \mathbbm{1}_{acc(\lambda_{z,y}) > 0.5}, & \text{if $z$ contained in $x$} \\
                0, & \text{otherwise}\\
            \end{cases}
\end{align}
Indeed, these are only two examples on how one can augment Nemo to support different IDP workflows in practice. We look forward to seeing more future work that explores other possibilities.
\vspace{-1mm}
\section{Conclusion}
We formalize and study the problem of IDP, on strategically selecting development data for more efficient LF development and exploiting the development context for better LF modeling. We introduce Nemo for IDP, which is built upon the novel SEU selection strategy and the contextualized learning pipeline. We validate that Nemo leads to more efficient and productive DP pipeline over the existing prevailing workflow, with an average 20\% (and up to 47\%) improvement across various datasets and tasks.


\clearpage

\balance
\bibliographystyle{ACM-Reference-Format}
\bibliography{main}

\end{document}